\pdfoutput=1
\documentclass[11pt]{article}

\usepackage{acl}

\usepackage{times}
\usepackage{latexsym}
\usepackage{graphicx}
\usepackage{float}
\usepackage{lineno}
\usepackage{multirow}
\usepackage{booktabs} 
\usepackage{adjustbox}
\usepackage{svg}
\usepackage{hyperref}
\usepackage{enumitem}
\usepackage{array}
\usepackage{tabularx}
\usepackage{amsmath}
\usepackage{hyperref} 

\usepackage[T1]{fontenc}
\usepackage[utf8]{inputenc}

\usepackage{microtype}

\usepackage{inconsolata}

\usepackage[T1]{fontenc}
\title{We Care: Multimodal Depression Detection and Knowledge Infused Mental Health Therapeutic Response Generation}

\author{Palash Moon \and Pushpak Bhattacharyya \\
  Department of Computer Science and Engineering, \\
  Indian Institute of Technology Bombay\\
  \{\texttt{palash, pb}\}@cse.iitb.ac.in}

\begin{document}
\maketitle
\begin{abstract}
The detection of depression through non-verbal cues has gained significant attention. Previous research predominantly centred on identifying depression within the confines of controlled laboratory environments, often with the supervision of psychologists or counsellors. Unfortunately, datasets generated in such controlled settings may struggle to account for individual behaviours in real-life situations. In response to this limitation, we present the Extended D-vlog dataset, encompassing a collection of $1,261$ YouTube vlogs. Additionally, the emergence of large language models (LLMs) like GPT3.5, and GPT4 has sparked interest in their potential they can act like mental health professionals. Yet, the readiness of these LLM models to be used in real-life settings is still a concern as they can give wrong responses that can harm the users. We introduce a virtual agent serving as an initial contact for mental health patients, offering Cognitive Behavioral Therapy (CBT)-based responses. It comprises two core functions: 1. Identifying depression in individuals, and 2. Delivering CBT-based therapeutic responses. Our Mistral model achieved impressive scores of $\textbf{70.1\%}$ and $\textbf{30.9\%}$ for distortion assessment and classification, along with a Bert score of $\textbf{88.7\%}$. Moreover, utilizing the TVLT model on our Multimodal Extended D-vlog Dataset yielded outstanding results, with an impressive F1-score of $\textbf{67.8}$\%.
\end{abstract}.
\section{Introduction}
Depression is a prevalent and significant medical condition. It hurts one's emotional state, thought processes, and behaviour. It manifests as persistent feelings of sadness and diminished interest in previously enjoyed activities. This condition can give rise to various emotional and physical challenges, affecting one's ability to perform effectively both at work and in personal life. Depression symptoms range from mild to severe and can include persistent sadness, loss of interest in once-enjoyable activities, appetite changes, sleep disturbances, fatigue, psychomotor changes, feelings of worthlessness, cognitive challenges, and, in severe cases, suicidal Thoughts. Symptom severity varies, requiring careful clinical evaluation for diagnosis and treatment.\\
\noindent
\textbf{Motivation}: According to the Statistics of the World Health Organisation (WHO) $3.8$\% of the world's population experience depression, including $5$\% of adults less than $60$ years of age ($4$\% of men and $6$\% of women) and $5.7$\% of adults above $60$ years of age. Approximately $280$ million people have depression which depression is $50$\% more common in women than men. Depression is $10$\% more in pregnant women and women who have just given birth \cite{evans-lacko2018}. If the depression is left untreated can lead to several serious outcomes such as suicide \cite{ghosh2022}.\footnote{\url{https://www.who.int/news-room/fact-sheets/detail/depression}}
% bridge the gap between mental health professionals and patients. 
Currently, there is a lack of mental health practitioners globally, with a ratio of $1:10000$ mental health professionals per patient. Our objective is to reduce the gap between patients and mental health professionals. We aim to achieve this by providing an automatic way of predicting depression and offering therapeutic responses to users, which can mitigate distress to some extent. \\
\noindent
\textbf{Virtual Agent:} In the realm of mental health support, the notion of therapy chatbots has intrigued both researchers and the public since the introduction of Eliza \cite{shum2018eliza} in the 1960s. Recent advancements in large language models (LLMs) like ChatGPT have further fueled this interest. However, concerns have been raised by mental health experts regarding the use of LLMs for therapy as the therapy provided may not be accurate. Despite this, many researchers have begun exploring LLMs as a means of providing mental health support \cite{sharma2023human}.\\
Our understanding of how LLMs behave in response to clients seeking mental health support remains limited. It is unclear under what circumstances LLMs prioritize certain behaviours, such as reflecting on client emotions or problem-solving, and to what extent \cite{chung2023challenges}, \cite{ma2023understanding}. Given the critical nature of mental health support, it is essential to comprehend LLM behaviour, as undesirable actions could have severe consequences for vulnerable clients. Additionally, identifying desirable and undesirable behaviours can inform the adoption and improvement of LLMs in mental health support.\\
\textbf{Conclusion:} We harness the capabilities of the Vision-Language TVLT \cite{tang2022tvlt} Transformer model, known for its state-of-the-art performance in tasks like video-captioning and multimodal sentiment analysis. Acting as an encoder-decoder, TVLT \cite{tang2022tvlt} processes raw video, audio, and text inputs to generate a comprehensive multimodal representation, beneficial for downstream tasks. By augmenting wav2vec2 \cite{baevski2020wav2vec} features with spectrograms for audio, we achieved an impressive accuracy of 67.8\% on the extended D-vlog dataset. Using the Mistral-7b Instruct-v.02 Language Model (LLM) with Chain-of-Thought prompting, we achieve a high Bert score of 88.7 which is a similarity score and generate Cognitive Behavioral Therapy (CBT)-based responses. Our methodology yields notable results: a 70.1 F1-score for detecting cognitive distortion and a 30.9 F1-score for multi-label classification, identifying ten types of cognitive distortion.\\
Our Contribution are:
\label{sec: contribution}
\begin{itemize}[nosep]
    \item Extended D-vlog dataset (\textbf{Original no. of videos: 961},\textbf{ Total videos} (after adding 300 videos to the Original dataset): \textbf{1261}) which contains videos of various types such as Major depressive disorder, postmortem disorder, anxiety and videos from different age group and gender which was lacking in the original D-vlog dataset. \hyperref[sec:dataset]{(Section 3)}
    \item TVLT \cite{tang2022tvlt} model for depression detection, which outperforms baseline models by \textbf{4.3\%} and establishes a new benchmark, on the Extended D-vlog dataset.
    \hyperref[sec:Result]{(see section 6)}
    \item Replacing spectrogram with the combination of spectrogram and  wav2vec2 \cite{baevski2020wav2vec} features which capture the vocal cues associated with depression more effectively than spectrogram, which further increases the accuracy by \textbf{2.2\%}  resulting in the final F1-score of \textbf{67.8} \%.
    \hyperref[sec:Result]{(see section 6)}
    \item To the best of our knowledge, this work is the first to propose a virtual agent that delivers therapeutic responses to users using LLM with Domain Knowledge as an External Knowledge base on Mental Health.
\end{itemize}

\section{Related Work}
With the rise in mental health conditions, there's growing interest in detecting depression. However, there's a shortage of datasets for this purpose, largely due to privacy concerns, limiting public availability. Among the few publicly accessible datasets, the DAIC-WOZ \cite{gratch2014distress} is notable, featuring clinical interviews in text, audio, and video formats, relying on self-reporting via the PHQ-8 questionnaire. Another dataset, the Pittsburgh dataset, primarily contains audio and video clinical interviews. Despite its small size of 189 samples, the DAIC-WOZ \cite{gratch2014distress} remains valuable for research. The AViD-Corpus, used in AVEC 2013 \cite{valstar2013avec} and 2014 \cite{valstar2014avec} competitions, includes video recordings of various activities with self-reporting conducted in the presence of mental health professionals. While these datasets provide insights into depression patterns, their assembly in controlled environments may not fully represent typical behaviours of depressed individuals.
\begin{table}[h]
\small % Set the font size to \footnotesize
\renewcommand{\arraystretch}{1.1} % Adjust vertical spacing between rows
\begin{tabular}{cccc}
\toprule
\textbf{dataset} & \textbf{Modality} & \textbf{\# Subjects} & \textbf{\# Samples} \\
\midrule
DAIC-WOZ & A+V+T & $189$ & $189$ \\
Pittsburg & A+V & $49$ & $130$ \\
AViD-Corpus & A+V & $292$ & $340$ \\
D-vlog & A+V & $816$ & $961$ \\
\midrule % Use \midrule instead of \hline
E-Dvlog & A+V+T & $1016$ & $1261$ \\
\bottomrule
\end{tabular}
\caption{Comparision of various Depression datasets with E-Dvlog (Extended D-vlog). Where A: Audio, V: Video, T: Text.}
\end{table}

The use of social media for depression detection is increasingly preferred over clinical interviews due to its ability to capture patients' authentic behaviour. Unlike supervised interviews, social media datasets reveal atypical behaviours exhibited in daily life. In recent years, depression detection using text from social media has been focused on \cite{fatima2019prediction}, \cite{burdisso2019text}, \cite{chiong2021combining}. Various approaches have emerged to detect depression using data from platforms like Twitter, Reddit, and Facebook, focusing on textual-based features such as linguistic characteristics. For instance, \cite{yang2018integrating} utilized text and tags from micro-blogs in China to extract behavioural features for depression detection. However, there's a growing need to explore video data and multimodal fusion for more comprehensive detection methods. \\
Multimodal fusion combines various modalities to predict outcomes, and it's increasingly used for depression detection. \cite{haque2018measuring} utilized 3D facial expressions and spoken language features to detect depression. \cite{yang2018integrating}  integrated text and video features, employing deep and shallow models for depression estimation. \cite{ortega2019multimodal} proposed an end-to-end deep neural network integrating speech, facial, and text features for emotional state estimation. Although previous studies have explored depression detection using multimodalities, the combination of Multimodal Transformer with wav2vec2 features and spectrograms remains unexplored despite its potential for superior results.\\
\textbf{Virtual Agents:}
In recent years with the increase in mental health problems, people have started taking emotional support from text-based platforms such as in \cite{eysenbach2004health}, \cite{de2014mental}, (talkelife. co). there is also a rise in Empathetic virtual agents \cite{saha2022shoulder}, which impart empathy in their responses by giving motivational responses and responses with hope and reflections which is seen as an important to uplift the spirit of an individual who is seeking support.
Additionally, efforts have been made to enhance the therapeutic value of these platforms by incorporating insights \cite{fitzpatrick2017delivering} encouraging exploration through open-ended questioning, and providing guidance and problem-solving techniques, all aimed at aiding users in their healing process.

\section{Datasets}
\label{sec:dataset}
The D-vlog dataset \cite{yoon2022d} is a collection of Depression vlogs of various people posted on YouTube. The  D-vlog \cite{yoon2022d} dataset has $961$ vlogs in total out of which $505$ are categorized as depressive vlogs and $465$ are categorized as Non-depressive vlogs. However, the D-vlog dataset \cite{yoon2022d} has some limitations, such as the dataset majorly having Major Depressive Disorder and lacking Other Disorder such as Bipolar Disorder, Postmortem Disorder, and Anxiety with depression. Which does make the dataset more generalized. So, we extended the D-vlog dataset \cite{yoon2022d} by adding  $300$ more vlogs to the D-vlog dataset \cite{yoon2022d} which now has more vlogs on various depressive disorders from varying age groups and different genders. Refer to this Figure \ref{fig:img1}. More data collection details are in Appendix \ref{sec:data_collection}. 

\subsection{Dataset Statistics:}
The extended D-vlog dataset has $1261$ vlogs with $680$ depressive vlogs and $590$ non-Depressive vlogs as Shown in below Table \ref{tab:Dataset_statistics}
\begin{table}[ht]
\centering
\normalsize
\renewcommand{\arraystretch}{1.1}
\begin{tabular}{cccc}
\hline
\textbf{} & \textbf{Gender} & \textbf{\# Samples} \\
\hline
\multirow{2}{*}{\textbf{Depression}} & Male & $273$ \\
& Female & $406$  \\
\hline
\multirow{2}{*}{\textbf{Non-Depression}} & Male & $232$  \\
& Female & $350$  \\
\hline
\end{tabular}

\caption{Extended D-vlog Statistics}
\label{tab:Dataset_statistics}
\end{table}

\begin{figure}[ht]
    \includegraphics[scale=0.65]{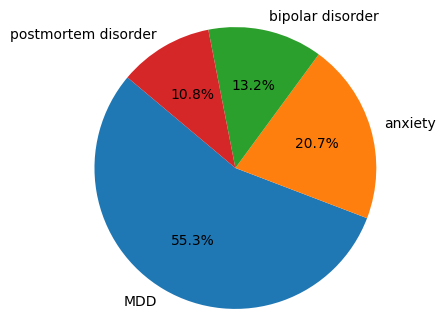}
    \caption{The Above figure shows the distribution of various types of Depressive vlogs. where MDD is Major Depressive Disorder, Bipolar Disorder is also called as Manic Disorder.}
    \label{fig:img1}
\end{figure}

\subsection{ Datasets for Therapeutic Conversations:}
Acquiring datasets of therapy conversations poses a significant challenge as they are typically private and rarely shared. Moreover, potential privacy issues may arise when exposing therapy datasets to public LLM APIs as they may contain sensitive client information.
Publicly available therapy conversation datasets are limited. Here, we use three datasets that carefully preprocess publicly available therapy. This ensures high-quality transcripts while maintaining the confidentiality of sensitive personal information. These datasets are 1. High-and-Low-Quality Therapy Conversation Dataset (High-Low Quality) \cite{perez2018analyzing} 2. HOPE Dataset \cite{malhotra2022speaker} 3. Motivate Dataset \cite{saha2022shoulder}. Further details can be seen in the Appendix section \ref{sec: data_theraputic}.

\section{Methodology}
The system is divided into two stages. \\
\textbf{Stage 1:} Detection of Depression where the video, audio and text are provided as input to the model for depression detection.\\
\textbf{Stage 2:} Provide a therapeutic response to the depressed user. The utterance that was given previously to detect depression. The same text utterance will be fed to the virtual assistant to find the type of distortion classification and after that, we generate the responses.
\subsection{Detection}
We use \textbf{TVLT} (Textless Vision Language Transformer) \cite{tang2022tvlt}, an end-to-end vision and language Multimodal transformer model that takes raw video, raw audio, and text as input to the transformer model. TVLT \cite{tang2022tvlt} is a textless model, which implicitly does not use text, but with the ASR model (whisper) \cite{radford2023robust}, we can extract text from the audio segments. The TVLT model is more effective for multimodal classification because the TVLT \cite{tang2022tvlt} model can capture visual and acoustic information, providing a more comprehensive fused representation of video, audio, and text.  \\
\begin{figure*}[ht]
    \centering
    \includegraphics[scale=0.50]{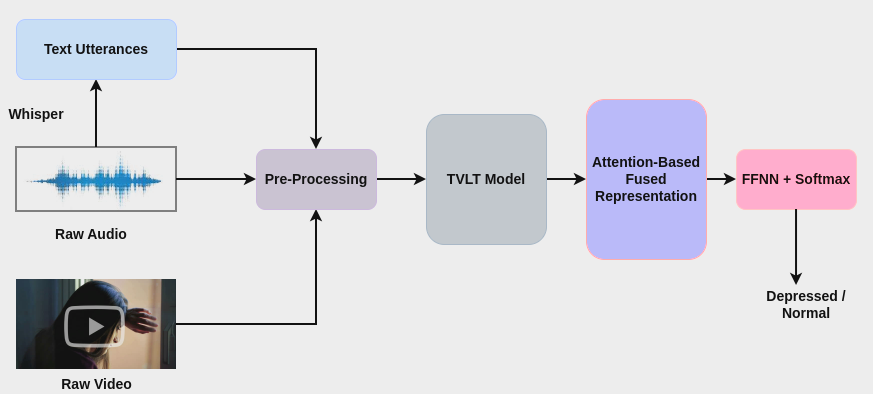}
    \caption{In the Above \textbf{Architecture} we leverage three different modalities such as video, audio and text where text is extracted from the audio segment using the Whisper ASR Model. we then preprocess all three modalities and pass them to the model where we get the fused representation of all three modalities. This fused representation is then passed to the feed-forward Neural Network with a sigmoid function to determine whether the individual exhibits signs of Depression or is in a Normal state.}
    \label{fig:img2}
\end{figure*}
\textbf{Textual Feature:} We make use of the powerful BERT \cite{kenton2019bert} Language model, a pre-trained model described to capture important features from text. This means we can understand not only the specific details in the text but also the overall context. These BERT embeddings help us understand text thoroughly, making them perfect for tasks like analyzing sentiment or identifying depression. We apply BERT \cite{kenton2019bert} to our text, using specific dimensions (dt = 786).\\
\textbf{Audio features: }We use a combination of techniques to analyze audio. Firstly, we generate spectrograms using the librosa library \cite{mcfee2015librosa} and extract low-level features. Additionally, we incorporate features from wav2vec2, which is described in \cite{baevski2020wav2vec}. The wav2vec2 features include various acoustic attributes such as MFCC \cite{5709752}, spectral \cite{pachet2007exploring}, temporal \cite{krishnamoorthy2011enhancement}, and prosody \cite{olwal2005interaction} features. These features help with identifying the pitch, intonation, and tempo of the audio segment. They are excellent at capturing both local and contextual information from the raw audio waveform. Finally, we compute the average across the spectrogram vector and the wav2vec2 vector to create our final audio representation. \\
\textbf{Video Features: }Our video processing pipeline involves several essential steps. First, we load the video file using a tool called VideoReader \cite{frith2005sound}. Next, we randomly select a subset of frames from the video clip. These frames are then resized and cropped to focus on the subject's frontal view. For extracting visual features, we rely on the powerful ViT (Vision Transformer) model introduced in \cite{dosovitskiy2020image}. This model helps us create what we call "vision embeddings." It does this by breaking down each video frame into smaller 16x16 patches. We then apply a linear projection layer to these patches, resulting in a $768$-dimensional patch embedding. This vision embedding module is a critical component of our model. It transforms each video frame or image into a sequence of $768$-dimensional vectors. These vectors are rich in both spatial and temporal information, making them invaluable for our model to comprehend the visual content within the input data.\\
We have implemented the architecture illustrated in Figure \ref{fig:img2}, where our TVLT \cite{tang2022tvlt} transformer model comprises a 12-layer encoder and an 8-layer decoder. To obtain the fused representation of all three modalities, We exclusively utilize the encoder portion of the model to generate fused representations for depression prediction tasks. Our evaluation, conducted on the extended D-vlog\cite{yoon2022d} dataset comprising $35,046$ video clips from $1016$ speakers, involves transcription using an ASR model with manual error correction. We split the data into a $7$:$1$:$2$ train-validation-test ratio and employ weighted accuracy (WA) and F1-score metrics. Additionally, we add task-specific heads on top of the encoder representation and train the model using binary cross-entropy loss for each downstream task.
\begin{align}
    L(y, \hat{y}) & = -\left[y \log(\hat{y}) + (1 - y) \log(1 - \hat{y})\right]
\end{align}
where y: True label and $\hat{y}$: Predicted label.

% updation
\subsection{Prompting with LLMs:}
During the discussion with the psychotherapist, we learned that identifying the ABCs is crucial for identifying distortions and determining the type of distortion. The ABCs stand for Activation Events, Beliefs, and Consequences. 
\begin{itemize}
    \item \textbf{Activation Event (A):}Identifying the specific situations or events that trigger emotional responses helps in pinpointing the cause of the distortion.
    \item \textbf{Beliefs (B):} which are the patient's thoughts and interpretations regarding the mentioned Activating Event.
    \item  \textbf{Consequences (C):}  The term refers to the impact that the Activating Event has had on an individual's life.
\end{itemize}

% new
Through the analysis of the ABCs, it becomes easier to understand the distortions and their underlying reasons \cite{dryden2012abcs}, \cite{lam2008cognitive}. Identifying these distortions and the reasons behind them can help challenge the distorted beliefs by asking why the individual is feeling that way, and reassuring them that these beliefs are normal. Ultimately, this can lead to a therapeutic response such as cognitive reconstruction. To determine whether ABC's generated are correct we performed a human evaluation on 200 samples. Additional information is provided in Appendix \ref{sec: human_eval}. After generating the ABCs, we input them along with an additional few shot prompts to the Mistral-7B-Instruct-v0.2 model \footnote{https://huggingface.co/mistralai/Mistral-7B-Instruct-v0.2}. By doing so, we determine whether the assessment exhibits cognitive distortion and identify the specific type of distortion present. This process enables us to offer the appropriate therapeutic response to the user based on the type of distortion identified.
We use the RAG \cite{lewis2020retrieval} pipeline incorporating domain-specific documents as an external knowledge base. This external knowledge is employed to validate and correct the responses generated and fine-tune the mistral model \cite{jiang2023mistral} which was fine-tuned on the motivate \cite{saha2022shoulder} and hope \cite{malhotra2022speaker} dataset. While generating responses to user queries, we utilize a system prompt as given in (Appendix \ref{sec:response_generation})

\section{Experiments}
\textbf{Detection:} To obtain a fused representation of audio, video and text modalities, we employ trained text-based TVLT \cite{tang2022tvlt} model on the video dataset and subsequently fine-tune on the extended D-vlog \cite{yoon2022d} dataset. We split our dataset into train, valid and test sets in the ratio of $7$:$1$:$2$. Details are in the Appendix \ref{sec: setup-detection}\\
\textbf{Distortion Identification:} We employ the "Mistral-7B-Instruct-v0.2" model to prompt and determine two things directly: firstly, whether an individual exhibits cognitive distortions based on provided context, and secondly, if so, to identify the specific type of cognitive distortion present on the extended D-vlog test dataset.
We use few-shot chain-of-thought \cite{wei2022chain} techniques to identify cognitive distortions, including ABC \cite{dryden2012abcs} prompts and pinpointing the distorted parts. Additionally, we explore providing reasoning for the distorted portions identified. Prompt details are given in the Appendix \ref{sec: identification_prompt}. Results can be seen in Table \ref{tab:mistral_table}.\\
\textbf{Response Generation:} We use pre-trained Mistral-7B models \cite{jiang2023mistral}, fine-tuned on hope \cite{malhotra2022speaker} and motivation data \cite{saha2022shoulder}, employing PEFT QloRA \cite{dettmers2024qlora}- a method that combines 4-bit quantization with low-rank adapters for improved memory usage and computational efficiency—to generate therapeutic responses. Additionally, we implemented a chain-of-thought \cite{wei2022chain} with an \cite{lewis2020retrieval} RAG pipeline to ensure accurate responses without generating false information, utilizing Adam's Optimizer with a learning rate set to $0.00025$, known for its superior results. we leverage the Mistral-7b as a Large language model, utilizing the pre-trained RAG model "thenlper/gte-large" from the Hugging Face library. The chunk size used here is $256$ and employs the vectorStoredIndex as an indexing mechanism for the storage and retrieval of embeddings from documents.

\section{Result and Discussion}
In this section, we will cover the results on the extended D-vlog Dataset \cite{yoon2022d}, the clinical Diac-woz dataset and results on distortion classification and response generation.
\label{sec:Result}
\subsection{Result on extended D-vlog dataset}
To analyse the importance of each modality for depression detection, we trained our model on each modality separately and reported the results in Table \ref{tab:table5} below.
\begin{table}[ht]
\centering
\begin{tabular}{|c|c|}
\hline
\textbf{Modalities} & \textbf{F1-scores}  \\
\hline
T & $0.57$  \\
A & $0.60$  \\
V & $0.56$  \\
\hline
V + A & $0.631$\\
V + T & $0.628$\\
A + T & $0.634$\\
\hline
V + A + T & \textbf{0.656}\\
\hline
\end{tabular}
\caption{Results obtained on the Extended D-vlogs dataset via experiments with Video (V), Audio (A), Textual (T).}
\label{tab:table5}
\end{table}
We discovered that the audio modality outperforms other modalities in terms of F1-Score, indicating its significance in depression detection. This suggests that individuals with depression exhibit distinct speech patterns. Although audio features outweigh visual ones, combining both modalities results in superior performance compared to using audio alone. Additionally, combining audio and text modalities surpasses using audio alone. Finally, incorporating all three modalities yields the best results, highlighting the effectiveness of considering audio, visual, and textual features and their interactions in depression detection.
\begin{table}[ht]
\centering
\begin{tabular}{|c|c|}
\hline
\textbf{Modalities} & \textbf{F1-scores}  \\
\hline
V + A + T & $0.656$\\
V + A + T(Mask) & $0.663$\\
V + A(W2V2+Spect) + T & \textbf{0.678}\\
V(Mask) + A + T & $0.661$\\
\hline
\end{tabular}
\caption{Results obtained on the Extended D-vlogs dataset via experiments with Video (V), Audio (A), Textual (T). T(Mask) is text with word-masking, V(Mask) is Video frames with frame-masking and A(W2V2+Spect) is Audio with wav2vec2 +spectrogram features.}
\label{tab:table6}
\end{table}

\begin{table*}[ht]
\centering
\large
\renewcommand{\arraystretch}{1.3}
\begin{tabular}{ccccc}
\hline
\textbf{Model Type} & \textbf{Model} & \textbf{Precision} & \textbf{Recall} & \textbf{F1-Score} \\
\hline
\multirow{3}{*}{\textbf{Fusion Baseline}} & Concat & $62.51$ & $63.21$ & $61.1$\\
                                          & Add & $59.11$ & $60.38$ & $58.1$ \\
                                          & Multiply & $63.48$ & $64.15$ & $63.09$ \\
\hline
\multirow{1}{*}{\textbf{Depression Detector}} & Cross-Attention & $65.4$ & $65.5$ & $65.4$\\
\hline
\multirow{1}{*}{\textbf{Our Model}} & TVLT Model & \textbf{67.3} & \textbf{68.3} & \textbf{67.8}\\
\hline
\end{tabular}
\caption{Comparison of various baseline models with our model on the extended D-vlog dataset}
\label{tab:table7}
\end{table*}
\noindent
Introducing random word masking in text significantly improves the model's textual understanding, resulting in a performance boost of $0.007$. Similarly, applying frame masking to video data, alongside audio and text modalities, enhances performance by $0.005$. These results highlight the effectiveness of incorporating diverse modalities. The table (Table \ref{tab:table7}) underscores the importance of leveraging text, video, and audio modalities with wav2vec2 \cite{baevski2020wav2vec} features and spectrograms, leading to an impressive F1-score of $67.8\%$.

\noindent
We extensively evaluated the TVLT \cite{tang2022tvlt} model's performance on the D-vlog \cite{yoon2022d} dataset, comparing it with several baseline models to gauge its effectiveness in depression detection. The TVLT \cite{tang2022tvlt} model outperformed the Cross Attention State-of-the-Art model by $2.2$\%, establishing itself as the new benchmark for the D-vlog \cite{yoon2022d} dataset. Its exceptional performance showcases its ability to understand the dataset's complexity, potentially inspiring further advancements in multi-modal analysis and deep learning techniques.

% Experiment on Clinical Dataset 
\subsection{Result on the clinical dataset: DAIC-WOZ}
We tested our proposed extended D-Vlog dataset for depression detection in the clinically labelled DAIC-WOZ dataset, using the same feature extraction process.
\begin{table}[ht]
\centering
\begin{tabular}{ccccc}
\toprule
\textbf{Train}&\textbf{Test}&\textbf{Precision}&\textbf{Recall}&\textbf{F1-score} \\
\midrule
DW&DV& $62.14$ & $62.38$ & $62.26$ \\
DV&DV& $66.40$ & $66.57$ & $66.48$  \\
DW&DW& $64.57$ & $54.63$ & $59.19$    \\
DV&DW& $69.45$ & $57.26$ & $62.77$   \\
\bottomrule
\end{tabular}
\caption{Results between extended D-Vlog
and DAIC-WOZ datasets. DV and DW denote D-Vlog and
DAIC-WOZ, respectively}
\label{tab:mytable}
\end{table}
We conducted four experiments with our model, including training and testing with extended D-Vlog, training with DAIC-WOZ and testing with extended D-Vlog, training and testing with DAIC-WOZ, and training with extended D-Vlog and testing with DAIC-WOZ. The results showed that the model trained with extended D-Vlog achieved better depression detection performance in both datasets. This suggests that D-Vlog's features, captured in daily life, are more useful than those in the DAIC-WOZ dataset, developed in a laboratory setting.

% Accuracies
\subsection{Results of cognitive distortion:}
\begin{table}[ht]
\centering
\begin{tabular}{p{4.5cm}p{1cm}p{1cm}}
\toprule
\textbf{Methods}&\textbf{DA F1-W}&\textbf{DC F1-W}\\
\midrule
Mistral & $62.4$ & $21.5$  \\
Mistral+FCOT & $63.9$ &  $22.3$   \\
Mistral+FCOT+ABC & $65.6$ & $27.8$  \\
Mistral+FCOT+ABCD & $67.3$ & $29.0$ \\
Mistral+FCOT+ABCDR & $\textbf{70.1}$ & $\textbf{30.9}$ \\
\midrule
ChatGPT + FCOT + ABC & $57.6$ & $20.4$  \\
ChatGPT + FCOT + ABCD & $59.1$ &  $21.0$  \\
ChatGPT + FCOT + ABCDR & $63.5$ &  $23.6$  \\
\bottomrule
\end{tabular}
\caption{DA: Distortion Assessment, DC: Distortion Classification, F1-W: F1-weighted, Fcot: Few-shot chain-of-thought, A: Activation Event, B: Belief, C: Consequences, D: Distored Part, RAG: Retrieval Augmented Generation}
\label{tab:mistral_table}
\end{table}
\noindent
We utilize the Mistral-7b \cite{jiang2023mistral} model to assess the F1-score for distortion assessment and classification across the ten types of distortion. Notably, we discover that integrating the ABC \cite{dryden2012abcs} framework from Cognitive Behavioral Therapy (CBT), which identifies Activation Event (A), Beliefs (B), and Consequences (C), notably enhances both the F1-score for distortion assessment and classification. Furthermore, by identifying the distorted segment within the context and providing reasoning behind its classification, we further enhance the F1-score for assessment and classification to \textbf{$70.1$} and \textbf{$30.9$}, respectively. We compared results from Instruct-Mistral-v.02 with ChatGPT and found Mistral performed well for distortion assessment and classification. This is because Mistral could identify distorted parts while ChatGPT couldn't discern them from context. \\
\\
\textbf{Ablation Study:} 
\begin{table}[ht]
\centering
\begin{tabular}{p{4.5cm}p{1cm}p{1cm}}
\toprule
\textbf{Methods}&\textbf{DA F1-W}&\textbf{DC F1-W}\\
\midrule
Mistral+FCOT+A & $51.9$ & $20.6$  \\
Mistral+FCOT+B & $\textbf{65.0}$ & $22.0$ \\
Mistral+FCOT+C & $63.4$ & $\textbf{23.5}$ \\

\bottomrule
\end{tabular}
\caption{DA: Distortion Assessment, DC: Distortion Classification, F1-W: F1-weighted, Fcot: Few-shot chain-of-thought, A: Activation Event, B: Belief, C: Consequences}
\label{tab:ablation}
\end{table}
We explore various settings to demonstrate the effectiveness of incorporating ABCs and see the impact of each on the assessment and classification. which shows that the beliefs and consequences are important measures for cognitive distortion. As we can see from the Table \ref{tab:ablation}.

\subsection{Results on Response generation:}
% this for the Acuuratness of the LLM
We have devised a prompt employing Cognitive Behavioral Therapy (CBT) techniques to craft therapeutic responses. Using the Mistral \cite{jiang2023mistral} prompt, we generate responses and compare them to the therapist's provided ground truth. Our analysis reveals a semantic similarity of $88.7$\% and $86.7$\% with Mistral \cite{jiang2023mistral} and LLama \cite{touvron2023llama} prompts respectively. This indicates that the generated responses closely align with the ground truth, affirming their semantic similarity.
\begin{table}[ht]
\centering
\begin{tabular}{p{1cm}p{1.2cm}p{1.5cm}p{2cm}}
\toprule
\textbf{Models}&\textbf{BLEU-4}&\textbf{ROUGE-L}&\textbf{BERT Score}\\
\midrule
Mistral & 25 & 23.5 & 88.7 \\
LLama & 21.6 & 18.8 & 86.7 \\
\bottomrule
\end{tabular}
\caption{Results on Bleu-4 score, Rouge-L score, and Bert score were evaluated for both the fine-tuned Mistral \cite{jiang2023mistral} model and the Lamma \cite{touvron2023llama} model.}
\label{tab:myresult}
\end{table}

\section{Qualitative Analysis}
Integrating wav2vec2 \cite{baevski2020wav2vec} features enhances our TVLT \cite{tang2022tvlt} model's depression detection accuracy by capturing vocal cues in audio data. This addition significantly improves performance compared to relying solely on spectrogram data, enabling our model to make more accurate predictions even in challenging scenarios.

Table \ref{tab:table8} highlights instances where our model accurately detects depression. In the first example, despite consistent facial expressions, the audio analysis reveals a monotone tone, low pitch, and crying, supported by distressing textual content. Our TVLT \cite{tang2022tvlt} model, augmented with wav2vec2 \cite{baevski2020wav2vec} and spectrogram features, accurately predicts depression, emphasizing wav2vec2's pivotal role in capturing vocal cues. 
In the second example, a girl's smile with tears indicates depression, detected accurately with wav2vec2 \cite{baevski2020wav2vec}. This showcases its ability to extract vital audio cues. In the third example, despite minimal facial expression variation and unremarkable audio, textual analysis reveals depression, posing a challenge to our model's accuracy.

\section{Summary, Conclusion \& Future Work}
In this study, we introduced an Extended D-vlog dataset with $1261$ videos, including vlogs by individuals with depression ($680$ videos) and those without ($590$ videos). Our goal is to detect depression in non-verbal and non-clinical vlogs using a TVLT \cite{tang2022tvlt} model, a multimodal transformer. We utilized text, video, and audio data, with visual embeddings from the Vit model and audio features from wav2vec2 \cite{baevski2020wav2vec} and spectrograms. The TVLT \cite{tang2022tvlt} model, incorporating all modalities, achieved an F1-score of $65.6$, which improved to $66.3$ with text word masking and $66.1$ with frame masking.
Our TVLT model, along with wav2vec2 \cite{baevski2020wav2vec} and spectrogram features, outperformed all baseline models on the D-vlog dataset and set a new benchmark on the Extended D-vlog dataset. We believe our dataset and model can play a crucial role in early depression identification through social media.  Our Mistral model achieved impressive F1 scores of $70.1$ and $30.9$ for distortion assessment and classification, along with a Bert score of $88.7$. In future, we are planning to create an LLM-based Psychologist Agent to converse with the user. 

\section{Limitation:} We have curated our dataset to exclusively feature vlogs from individuals who have experienced or are currently experiencing depression. We acknowledge the potential for bias inherent in this selection process. However, we have taken measures to mitigate this bias to the best of our ability. Our model encounters challenges in accurately predicting certain depressed classes, notably in cases such as 'smiling depression,' where individuals conceal genuine emotions behind a facade of cheerfulness and high functionality, making detection of this issue particularly challenging.

\section{Ethics Statement:}
All vlogs included in the dataset were voluntarily uploaded by individuals onto the YouTube platform, and each vlog is authored by its respective uploader. None of the vlogs in the dataset have been sourced from other platforms without explicit consent. All the data in the dataset has been sourced from open-access platforms, and none of the videos or text within it contain any offensive or derogatory language aimed at any particular team or entity.

\bibliography{custom}

\appendix

\appendix
\section{Appendix}
\subsection{TVLT Model:}
Textless Vision-Language Transformer (TVLT), is a model designed for vision-and-language representation learning using raw visual and audio inputs. Unlike traditional approaches, TVLT employs homogeneous transformer blocks with minimal modality-specific design and does not rely on text-specific modules such as tokenization or automatic speech recognition (ASR). TVLT is trained using masked autoencoding to reconstruct masked patches of continuous video frames and audio spectrograms, as well as contrastive modelling to align video and audio. Experiments demonstrate that TVLT achieves comparable performance to text-based models across various multimodal tasks, including visual question answering, image retrieval, video retrieval, and multimodal sentiment analysis. Additionally, TVLT offers significantly faster inference speed (28x) and requires only one-third of the parameters. These results suggest the feasibility of learning compact and efficient visual-linguistic representations directly from low-level visual and audio signals, without relying on pre-existing text data.
\subsection{Pretaining details:}
\begin{itemize}
    \item \textbf{HowTo100M:} We used HowTo100M, a dataset containing 136M video clips of a total of
134,472 hours from 1.22M YouTube videos to pre-trained our model. Our vanilla TVLT is pre-trained
directly using the frame and audio stream of the video clips. Our text-based TVLT is trained using the
frame and caption stream of the video. The captions are automatically generated ASR provided in the
dataset. We used 0.92M videos for pretraining, as some links to the videos were invalid to download.

\item \textbf{YTTemporal180M:} YTTemporal180M includes 180M video segments from 6M YouTube
videos that spans multiple domains, and topics, including instructional videos from HowTo100M,
lifestyle vlogs of everyday events from the VLOG dataset, and YouTube’s auto-suggested videos
for popular topics like ‘science’ or ‘home improvement’. 
\end{itemize}

\section{Dataset Collection:}
\label{sec:data_collection}
We have collected the dataset vlogs using certain keywords using YouTube API \cite{leon_yin_2018_1414418} and downloaded them using the yt-dlp package \footnote{\url{https://github.com/yt-dlp/yt-dlp/wiki/Installation}}.
\newline
\textbf{Depressive vlogs:} ‘depression daily vlog’, ‘depression journey’, ‘depression vlog’, ‘depression episode vlog’, ‘depression video diary’, ‘my depression diary’, and ‘my depression story’, 'postpartum depression vlogs', 'Anxiety vlogs'.
\newline
\textbf{Non-Depressive vlogs:} ‘daily vlog’, ‘grwm (get ready with me) vlog’, ‘haul vlog’, ‘how to vlog’, ‘day of vlog’, ‘talking vlog’, etc.
\newline
We used the same approach to collect the dataset as used in the D-vlog dataset \cite{yoon2022d}. We focused our analysis on vlogs featuring content creators who have a documented history of depression, currently manifesting symptoms of the condition. We specifically excluded vlogs that solely discussed having a bad day without a deeper connection to depressive experiences. 

\subsection{Datasets for Therapeutic Conversations: }
\label{sec: data_theraputic}
\begin{enumerate}
    \item \textbf{High-and-Low-Quality Therapy Conversation
    Dataset (High-Low Quality):} The initial dataset, established by \cite{perez2018analyzing}, encompasses 259 therapy dialogues, predominantly centring on evidence-based motivational interviewing (MI) therapy. Assessing the conversations by MI psychotherapy principles, the authors identify 155 transcripts of high quality and 104 of low quality within the dataset. Both high-quality and low-quality therapy dialogues conducted by human therapists are utilized to examine desirable and undesirable conversational behaviours.
    \item \textbf{HOPE Dataset:} The second dataset from \cite{malhotra2022speaker} was used to study dialogue acts in
    therapy. This dataset contains 212 therapy transcripts and includes conversations employing different types of therapy techniques (e.g., MI, Cognitive Behavioral
    Therapy).
    \item \textbf{MotiVAte Dataset:} The MotiVAte Dataset \cite{saha2022shoulder} contains 7076 dyadic conversations with support seekers who have one of the four mental disorders: MDD, OCD, Anxiety or PTSD. 
\end{enumerate}

\subsection{Experiments setup: Detection}
\label{sec: setup-detection}
\begin{table}[H]
\centering
\normalsize
\renewcommand{\arraystretch}{1.3}
\begin{tabular}{cccc}
\hline
\textbf{Gender} & \textbf{Train} & \textbf{Valid} & \textbf{Test}\\
\hline
\textbf{Male} & $354$ & $51$ & $100$ \\
\textbf{Female} & $530$ & $74$ & $152$ \\
\hline
\end{tabular}
\caption{Number of vlogs in Train, Valid and Test Split of Extended D-vlog dataset}
\end{table}
For training the model, we utilized Adam's Optimizer with learning rates ranging from $0.0001$ to $0.00001$ and batch sizes of $32$ and $64$. The model underwent four iterations with different seed values, each taking approximately three hours to train on the Nvidia RTX A6000. Binary cross-entropy served as our chosen loss function for the depression detection task, and F1 scores were reported based on the test set results.

The extended D-vlog dataset exhibits more representation of Female vlogs as compared to Male vlogs within the Depressed category, reflecting a high prevalence of depression among Females. In the non-depressive category similar trend is observed with more female representation than male vlogs as predominantly "get ready with me vlogs", and "Haul vlogs" are uploaded by females.

\section{Prompt Details}
\subsection{Response Generation Prompt}
\label{sec:response_generation}
Act like a mental health therapist skilled in Cognitive Behavior Therapy (CBT). Your client presents a cognitive distortion, and your task is to guide them towards healthier thinking. Your response should involve three key steps:
1. Reflective Inquiry: Acknowledge the distortion without judgment, exploring it with empathy and understanding.
2. Challenging Thoughts: Gently question the distorted thinking, uncovering its roots and promoting alternative perspectives.
3. Cognitive Restructuring: Offer practical strategies for reframing thoughts and fostering self-compassion, empowering your client to reshape their mindset.

\subsection{Prompt for Identification of Distortion}
\label{sec: identification_prompt}
you are a mental health therapist who uses Cognitive Behavioral Therapy (CBT) to give responses. Understand the following definitions:
Activating Event: This represents the specific situations or events that trigger emotional responses.
Beliefs: These are the patient's thoughts and interpretations regarding the mentioned Activating Event.
Consequences: What effect has happened due to the Activating Event on a person's life?
Cognitive Distortion: A cognitive distortion is an exaggerated or irrational thought pattern involved in the onset or perpetuation of psychopathological states, such as depression and anxiety.
Your task is to use Cognitive Behavioral Therapy (CBT), analyze the given question to identify the Activating Event, 
Belief, Consequences, Distortion Part in the Question. Follow the steps below: 
1. Identify the Activating Event: Pinpoint the specific situation triggering emotional responses.
2. Explore the Belief: Examine underlying thoughts, distinguishing between Rational and Irrational beliefs. Tell if it has Rational Belief or Irrational Belief. 
3. Assess the Consequences: Evaluate emotional, behavioural, and physiological outcomes resulting from beliefs.
4. Identify the Distorted part or sentence from the Question itself if present else none.
5. Using Question, Activation Event, Belief, Consequences and Distorted Part identify the Cognitive Distortion category from the above types if present else indicate none.  
6. Give a reason why you choose for a particular Cognitive Distortion and why not for the other Cognitive Distortion.
6. Provide an Assessment: if the case of Cognitive distortion provides yes else no.  
The Assessment should be "yes" or "no" only. \\
Followed by the types of cognitive distortion taken from \cite{shreevastava2021detecting}

\section{Human Evaluation}
\label{sec: human_eval}
We used only one human evaluator who has an idea about cognitive distortion and its types and we have shared with him 200 sample forms for the evaluation of the correctness of A: Activation event, B: beliefs, C: Consequences, D: Distorted part. The Percentage tells what percentage of time the model has given the correct value of the activation event, Beliefs, Consequences and distortion.
\begin{table}[h]
\centering
\begin{tabular}{p{3cm}p{3cm}}
\toprule
\textbf{Models} & \textbf{Percentage}\\
\midrule
Activation Event & 68\% \\
Beliefs & 52.5\% \\
Consequences & 63.2\%\\
Distorted & 41.2\% \\
\bottomrule
\end{tabular}
\caption{}
\label{tab:ABC}
\end{table}

\begin{table*}[t]
\centering
\begin{tabularx}{0.95\linewidth}{|X|p{2cm}|p{2cm}|p{2cm}|c|}
\hline
\textbf{Utterance} & \textbf{Ground Truth} & \textbf{Prediction (w2v2 + spect)} & \textbf{Prediction w/o (w2v2 + spect)} & \textbf{Video frames} \\
\hline
\centering I knew what I was feeling, but I don't think I was able to communicate entirely what I was feeling. Like I knew I had this pittish feeling in my stomach. I knew that I'd be scared to wake up. I didn't want to wake up. Yeah, I think waking up was tough because I didn't want to face a day. & Depression & Depression & Normal & \raisebox{-\totalheight}{\includegraphics[width=2cm, height=2.5cm]{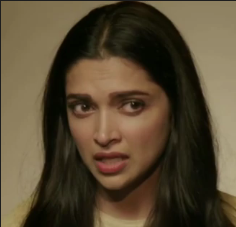}}\\
\hline
\centering Some days, it's hard to just move. It's... I like it. I, yeah, it's hard to get out of bed. It's hard to even go downstairs to get something to eat. & Depression & Depression & Normal & \raisebox{-\totalheight}{\includegraphics[width=2cm,height=2.3cm]{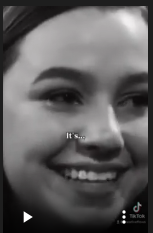}}\\
\hline
\centering  No concept of time, no sense of feeling. Have I become cold, dead to the world, where I once mattered? I can't even remember when I was important to someone last. Everything has escaped me. Deeper I fall into a void. & Depression & Normal & Normal & \raisebox{-\totalheight}{\includegraphics[width=2cm,height=2.3cm]{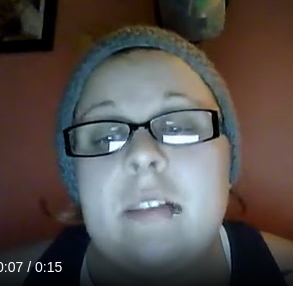}}\\
\hline
\end{tabularx}
\caption{A \textbf{Qualitative analysis}, In the given instances, the model, equipped with both wav2vec2 and spectrogram features, effectively detects depression through audio analysis. In the first example, despite seemingly normal facial expressions, the model accurately detects depression. In the second case, the model succeeds in identifying depression even when the individual smiles while crying, whereas the model relying solely on spectrogram data falls short in these situations. In the third scenario, the woman's facial expressions and audio do not exhibit evident signs of depression, while text analysis reveals potential indicators that challenge our model's accuracy, resulting in an incorrect prediction.}
\label{tab:table8}
\end{table*}
\end{document}